# Task-Oriented Robot-Human Handovers on Legged Manipulators *

Andreea Tulbure[†]
ETH Zürich
Zürich Switzerland
tulbure@mavt.ethz.ch

Carmen Scheidemann[†]
ETH Zürich
Zürich Switzerland
carmensc@ethz.ch

Elias Steiner
ETH Zürich
Zürich Switzerland

Marco Hutter
ETH Zürich
Zürich Switzerland

## Abstract

Task-oriented handovers (TOH) are fundamental to effective human-robot collaboration, requiring robots to present objects in a way that supports the human's intended post-handover use. Existing approaches are typically based on object- or task-specific affordances, but their ability to generalize to novel scenarios is limited. To address this gap, we present AFT-Handover, a framework that integrates large language model (LLM)-driven affordance reasoning with efficient texture-based affordance transfer to achieve zero-shot, generalizable TOH. Given a novel object-task pair, the method retrieves a proxy exemplar from a database, establishes part-level correspondences via LLM reasoning, and texturizes affordances for feature-based point cloud transfer. We evaluate AFT-Handover across diverse task-object pairs, showing improved handover success rates and stronger generalization compared to baselines. In a comparative user study, our framework is significantly preferred over the current state-of-the-art, effectively reducing human regrasping before tool use. Finally, we demonstrate TOH on legged manipulators, highlighting the potential of our framework for real-world robot-human handovers.

**ACM Reference Format:**
Andreea Tulbure, Carmen Scheidemann, Elias Steiner, and Marco Hutter. 2026. Task-Oriented Robot-Human Handovers on Legged Manipulators . In *Proceedings of the 21st ACM/IEEE International Conference on Human-Robot Interaction (HRI '26), March 16–19, 2026, Edinburgh, Scotland, UK*. ACM, New York, NY, USA, 9 pages. https://doi.org/10.1145/3757279.3785629

*This research was supported by ETH Zurich and has received funding from armasuisse Science and Technology. This work has been conducted as part of ANYmal Research, a community to advance legged robotics.

[†]Equal contribution

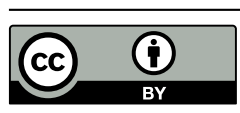

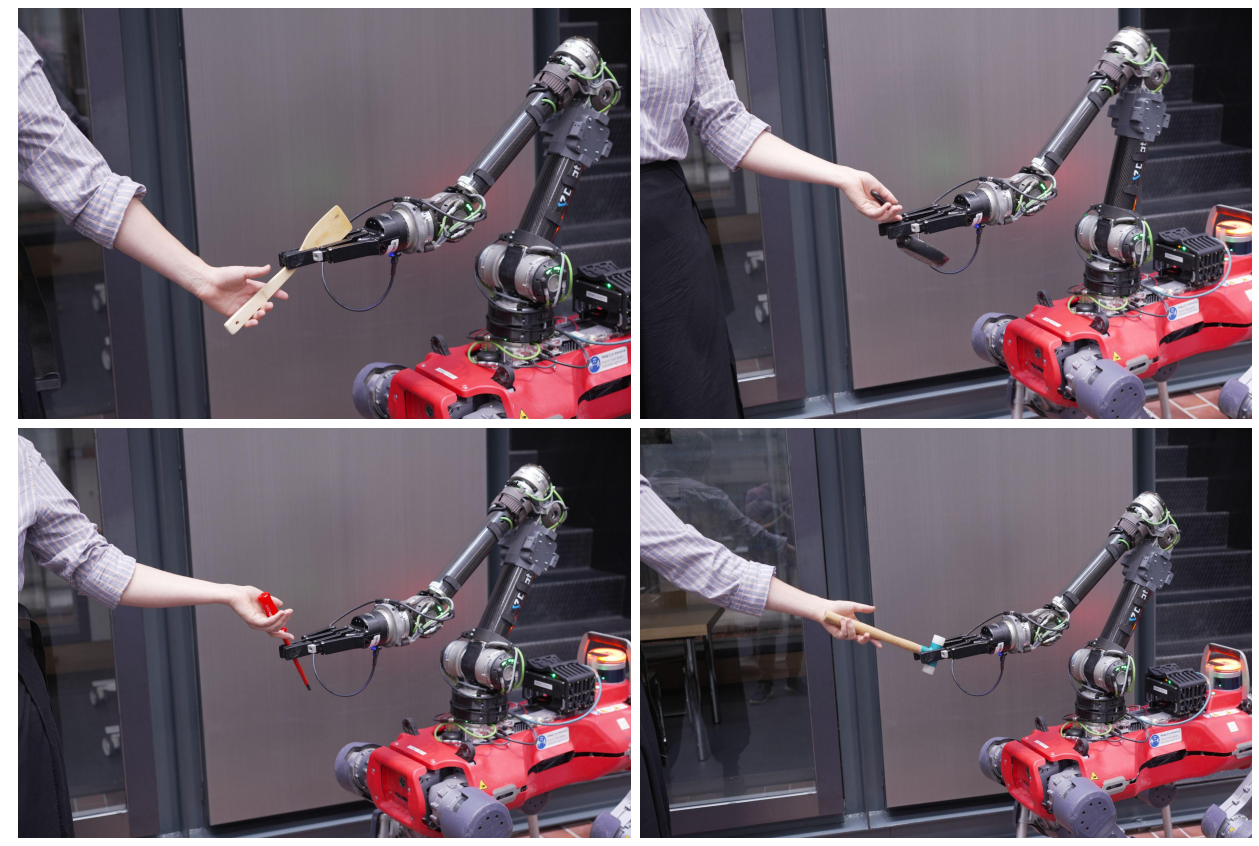

Figure 1: Our legged manipulator performing task-oriented robot-human handovers. Objects are oriented to support the human's intended use: spoon for stirring, pan for cooking, screwdriver for screwing, and hammer for hammering.

## 1 Introduction

As robots increasingly operate in everyday settings, seamless collaboration with people becomes a core requirement and object handovers are a key component of this interaction. Humans naturally perform task-oriented handovers, presenting objects according to the recipient's intended use to minimize regrasping (Fig. 1). In contrast, most robotic handovers remain task-agnostic, focusing on safe transfer rather than immediate usability [27].

The demands are even more pronounced for mobile legged manipulators. Their capability to traverse uneven terrain, fetch and deliver objects, and reposition relative to humans makes them particularly suitable for handover interactions in unstructured environments. However, these platforms introduce additional challenges: viewpoint shifts, motion blur, self-occlusions, and limited onboard computation and sensing. As a result, TOH on such platforms remains unexplored. Only one prior work studies handovers on legged manipulators [35], focusing solely on human-to-robot transfer and task-agnostic settings.

Achieving TOH requires reasoning not only about stable grasps, but also about the human's post-handover task. Prior work grounds handovers in affordances [2, 3, 7, 22, 26, 29, 38], linking grasping and functional use. However, affordances are often defined in a task- or object-specific manner, limiting generalization. We argue that a promising and efficient path is to treat TOH as an affordance transfer problem. Many affordances generalize across object classes (e.g., handles or blades), despite geometric differences. Leveraging these cross-category similarities enables robots to generalize to unseen objects without relearning affordances.

Affordance transfer has been explored across several domains: visual affordance prediction [13, 24, 39, 40], geometric or functional substitution [1, 10], and demonstration-retrieval methods [18, 20]. However, these methods struggle with large geometric variation, rely on brittle heuristics, or require extensive databases. Furthermore, few methods leverage the semantic reasoning of LLM/VLMs to align affordances across semantically distinct objects [8, 32, 34].

To address this gap and achieve zero-shot TOH, we propose *AFT-Handover*. It combines LLM-driven affordance reasoning with texture-based transfer. Given a novel object with no prior annotations, our framework: 1) uses an LLM to retrieve a proxy object from a small database based on task-specific functional similarity; 2) establishes part-level correspondences through LLM reasoning (e.g., knife blade ↔ screwdriver shaft, handle ↔ handle); 3) transfers affordances via 3D texture transfer, projecting annotated affordance fields from the proxy onto the novel object. This formulation treats affordances as continuous scalar textures over the object surface, enabling smooth interpolation and making the transfer robust to noise or partial views. The texture-transfer network is smaller, faster, and more data-efficient than affordance transfer networks, learning geometric correspondences rather than affordance classifiers - an advantage for handover tasks, especially with limited onboard compute on a mobile system.

By explicitly modelling affordance similarity across object classes and grounding them in the human's post-handover task, AFT-Handover bridges the gap between affordance transfer and TOH. The integration of LLM-driven commonsense reasoning with texture-based geometric affordance propagation introduces a conceptually and architecturally novel framework that supports reliable, task-aware affordance transfer across semantically and geometrically diverse objects. Overall, our contributions are:

- LLM-guided affordance reasoning. We introduce the first framework that leverages LLMs to reason about affordance similarity and establish part-level mappings for TOH, across both semantically similar and distinct object-task pairs.
- Zero-shot, inter-class transfer. We formulate affordance transfer as a texture transfer problem, enabling robust cross-category transfer even for geometrically dissimilar tools.
- TOH in real-world settings. We validate AFT-Handover on diverse objects and demonstrate TOH on legged manipulators, supporting the relevance for assistive HRI applications.
- System-level integration. AFT-Handover unifies perception, high-level semantic reasoning, geometric affordance transfer, and grasp planning into an end-to-end framework directly deployable on both fixed base and legged manipulators.

## 2 Related Work

In the following, we review the two strands of research that form the foundation of our work: TOH, which highlights the need to model human post-handover tool use in robot-human collaboration, and affordance transfer, which provides mechanisms to generalize functional knowledge across objects.

### 2.1 Task-oriented handovers

Most prior works on task-oriented robot-human handovers are based on grasp affordances [2, 3, 7, 22, 26, 29, 38]. Aleotti et al. [2] and Ortenzi et al. [26] manually avoid human grasp affordances. Demonstrating improved generalization capabilities, Ardon et al. [3] jointly optimize affordances, task objectives, and human mobility constraints, and show that clustering objects by shape enables generalization to unseen yet semantically similar tools. Meng et al. [22] predict human grasp affordances without modelling post-handover tasks and use heuristics to place the robot's grasp opposite the predicted human area. To move beyond such heuristics, Wang et al. [38] predict 3D human contact affordance maps, considering only object geometry, without task context. Robot grasps candidates are reranked to avoid the predicted human contact regions. Their evaluation, however, is simulation-only, and generalization is limited by dataset size and diversity. While grounded in affordances, these TOH approaches often generalize poorly since affordances are modelled per object or task, restricting inter-class transfer.

To address this limitation, Meiying et al. [29] use tool-use demonstrations to capture object geometry and task context information. This allows the robot to adapt handover configurations across various scenarios, but generalization across diverse objects remains inconclusive. LLM-Handover [36] further improves zero-shot reasoning for novel object-task pairs, including unconventional ones (e.g., using a screwdriver to hammer) by exploiting the semantic reasoning capabilities of LLMs. However, results are sensitive to task specification and part segmentation, and performance degrades for unconventional or less semantically aligned tasks.

### 2.2 Affordance transfer

Affordance transfer aims to generalize functional knowledge across objects, allowing robots to reason about novel tools by analogy to familiar ones. Visual affordance-transfer methods [13, 24, 39, 40] learn actionable regions from visual cues but degrade under large geometric variation. Affordance Transfer Learning (ATL) [16] demonstrates transfer of affordances across object categories to improve zero-shot generalization for human-object interaction detection. Another line of work studies functional substitution: finding alternative tools with geometries that support the same actions [1, 10, 30].

Retrieval-based methods extend affordance transfer by exploiting large demonstration databases. Robo-ABC [18] generalizes affordances by retrieving functional correspondences from a memory built from human demonstrations and applying them to novel objects. Similarly, RAM [20] combines retrieval with multimodal embeddings to condition affordance transfer on task descriptions and object observations. Finally, some works incorporate the reasoning capabilities of LLM and vision-language models (VLM) into affordance learning. AffordGrasp [34] uses VLMs for open-vocabulary, task-oriented grasping, while RTAGrasp [8] transfers affordance knowledge from human videos by aligning task semantics through language. These works underline the potential of language-guided reasoning for affordance transfer.

Task-oriented grasping (TOG) can be seen as affordance transfer, since predicting grasps for new tasks requires identifying task-relevant affordance regions. TOG-Net [9] and TaskGrasp [25] show that encoding task semantics and constraints allows generalization to novel objects and tasks. OS-TOG [15] performs one-shot TOG by reusing affordance knowledge from exemplars of the same object class, while the TD-TOG [14] dataset benchmarks zero- and one-shot TOG generalization. More recently, LLMs and VLMs have been exploited to extend this reasoning to open-vocabulary settings [8, 23, 32–34]. While these works hint at the potential for handover applications, they primarily optimize the robot grasp and do not consider the human's intended use after the handover.

However, visual methods often struggle when object geometries diverge significantly, tool substitution approaches rely on geometric heuristics, and retrieval-based methods require large demonstration databases. TOG methods demonstrate affordance-centric generalization, but primarily focus on grasps rather than affordance maps.

## 2.3 Summary

Overall, despite this progress, current TOH approaches either lack generalization or depend on accurate part detection, while most affordance-transfer methods do not incorporate the human task. Our framework addresses these gaps through LLM-guided semantic reasoning combined with a novel representation of affordances as continuous 3D textures. This translates affordance transfer into a texture transfer problem, enabling smoother and more data-efficient transfer via a lightweight texture-transfer network rather than a heavy affordance-prediction model. Unlike OS-TOG [15], which transfers affordances only within the same object class, and LLM-Handover [36], which provides semantic reasoning but no inter-class affordance transfer, our method unifies part-level semantic alignment with cross-category, task- and object-conditioned affordance transfer to enable robust, zero-shot TOH.

# 3 Methodology

## 3.1 Problem Formulation and Nomenclature

We consider a fixed or mobile base robot equipped with a two-finger parallel-jaw gripper. The framework input is (i) an RGB-D image of the scene and (ii) a natural language description $\mathcal{T}$ specifying the tool and its intended post-handover task. The goal is to predict a single, task-compliant grasp pose $g^* \in SE(3)$ for the robot, suitable for TOH. We define the *canonical orientation* as the normalized object point cloud (centering, scaling, and axis alignment) for meaningful cross-object comparison. The *query object* denotes the tool the human intends to use.

## 3.2 AFT-Handover

Our proposed framework, AFT-Handover, couples LLM-guided task reasoning with a texturization-based affordance transfer network to enable generalizable TOH. The pipeline, illustrated in Fig. 2, integrates perception, reasoning, planning, and affordance transfer and produces grasp poses optimized for both stable robot execution and the human's intended post-handover task. For Object Segmentation and Grasp Generation, we utilize existing open-source frameworks, namely LangSAM [21], a vision-language segmentation model that supports open-vocabulary queries, and Edge Grasp [17]. Grasp execution is handled by a motion planner and low-level controller specific to each robot. Next, we detail the Database, High-level LLM-Reasoning, Affordance Transfer, and Grasp Selection modules.

### 3.2.1 *Database.*

Humans rarely learn affordances from scratch for new tools. Instead, they reason by analogy: a previously unseen object is interpreted in relation to known ones. Following this intuition, our method leverages a database to retrieve an object prior that matches in terms of geometry and task semantics, enabling the robot to interpret new objects by analogy instead of re-learning affordances. Each entry of the database consists of a representative tool (e.g., mug, hammer, pan), its associated task (e.g., drinking, hammering, cooking), and a set of annotated subparts (e.g., handle, head, blade) together with their corresponding affordance annotations in the form of heatmaps on reference point clouds.

The reference affordances are extracted from the Contact-DB dataset [5], which was built by asking human participants to grasp 3D-printed household objects coated with thermochromic paint. This process captures fine-grained maps of where hands naturally make contact, delivering 3D object meshes annotated with human contact regions. We average the class-specific contact data across all users and threshold into binary grasp/non-grasp regions to achieve a single representation of the approximate human-occupied area.

Given a novel object-task pair, the LLM retrieves the most relevant exemplar from the database and establishes part-level correspondences (e.g., knife blade ↔ screwdriver shaft). In this way, the database acts as the bridge between high-level task semantics (captured by the LLM) and low-level affordance transfer.

### 3.2.2 *High-level LLM-Reasoning.*

This module operates in two steps: *(i) Task reasoning:* Given the query tool and post-handover task, the LLM outputs a textual description of how the task should be performed, identifying which part should be grasped and which part must remain free for task execution. This allows reasoning beyond strictly geometric cues, such as handing over a screwdriver by the shaft so the handle remains unobstructed. *(ii) Part localization:* To meaningfully align a query object with its matching database entry, we need approximate positions of its key functional parts. Precise 3D segmentation is not required; rough estimates are sufficient to orient the object consistently. Using the segmentation mask, we crop the RGB image around the tool and compute its principal axis via PCA. We then divide the object along this axis into coarse regions. Each region is presented to the LLM, which assigns semantic labels (e.g., "handle end", "blade tip"). These labelled regions are back-projected into 3D, yielding rough part centers. While approximate, these centers provide the anchors necessary for the canonical alignment step in Sec. 3.2.3, making the affordance transfer robust to rotational variations.

### 3.2.3 *Affordance Transfer.*

As illustrated in Fig. 3, the affordance transfer happens in three stages. In *(i) Functional Part Matching*, the LLM selects the most similar database object-task pair and generates a mapping between query and database entry subparts (e.g., pan handle ↔ mug handle). With *(ii) Query Point Cloud Alignment*, we address the problem that raw point clouds obtained from RGB-D sensors are inconsistent: they often capture only a partial view of the object, at arbitrary scale and rotation. Since our affordance transfer network is not inherently rotation- or scale-invariant, misalignment directly degrades descriptor matching and prevents reliable affordance transfer. To handle this, we normalize the query point cloud by centering, scaling, and rotating it into a canonical orientation before aligning it with the database entry. This ensures that part-level correspondences remain comparable across objects, despite differences in viewpoints. For alignment, we use the identified part centers: The alignment is performed such that the axes formed by the centers of the relevant identified parts of both tools are aligned. For *(iii) Affordance Transfer*, we apply a point-to-point texture transfer network [6], which learns self-supervised correspondences between non-rigid 3D shapes across modalities (e.g., textured meshes and raw point clouds). The network learns to predict functional maps between shapes in an unsupervised manner, capturing dense correspondences without requiring ground-truth annotations. During inference, the affordance annotations from the selected database object are transferred onto the aligned query point cloud by embedding the inputs into the descriptor space and establishing correspondences by nearest-neighbor search. The result is a color-coded point cloud (red = human grasp area, blue = free) that guides grasp selection.

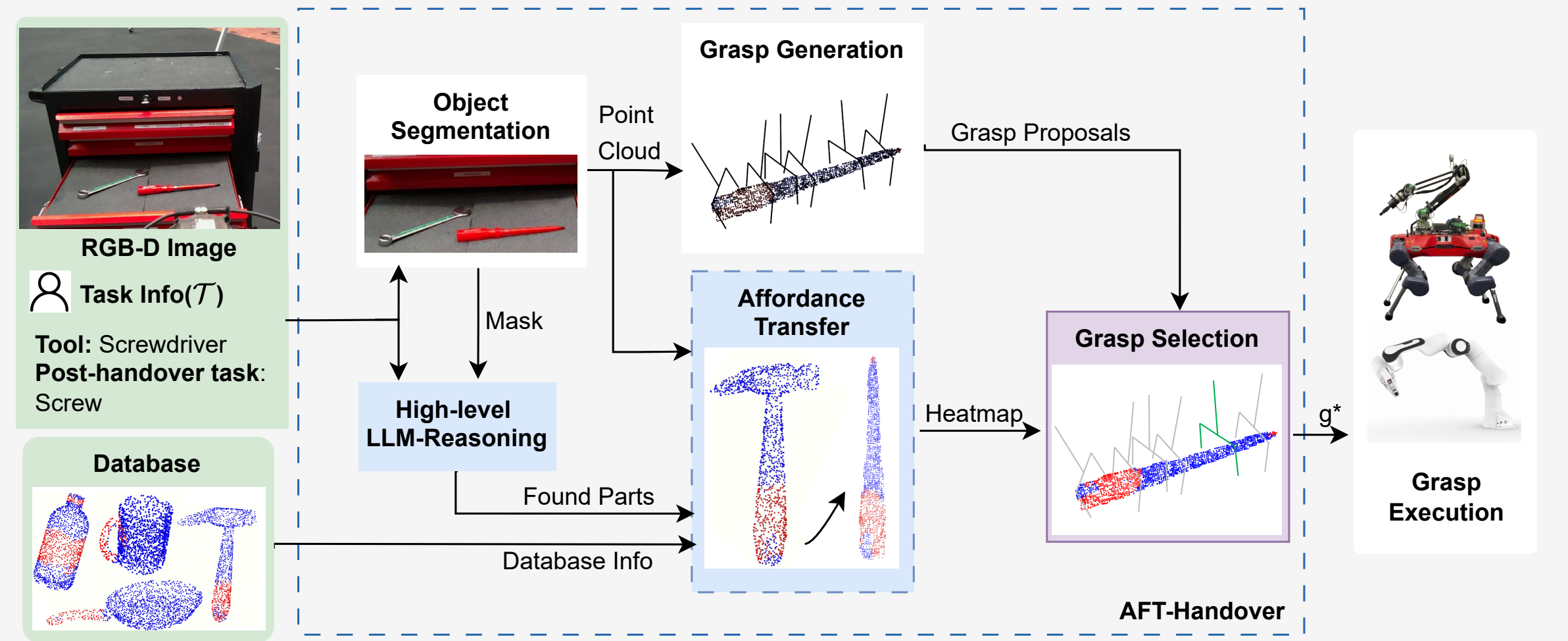

Figure 2: **Overview of AFT-Handover**. The pipeline integrates perception (Object Segmentation), grasping (Grasp Generation), reasoning (High-level LLM-Reasoning), and affordance transfer (Affordance Transfer) to select a task-compliant grasp pose $g^*$ in Grasp Selection. This grasp is executed by the robot. Proposed components are highlighted with colored boxes, existing modules are white (●). Novel components using LLMs are colored blue (●), new components without LLMs are purple (●), and the inputs to the framework are marked in green (●).

*3.2.4 Grasp Selection.* Candidate grasps are scored using a normalized objective that balances semantic compliance, ergonomic suitability, and feasibility:

$$s = w_{heat} \cdot \tilde{s}_{heat} + w_{dist} \cdot \tilde{d}_{human} + w_{feas} \cdot \tilde{s}_{feas}, \tag{1}$$

where $\tilde{s}_{heat} \in [0, 1]$ is the fraction of predicted human-grasp area points touched by the candidate grasp; $\tilde{d}_{human} \in [0, 1]$ is the distance between the grasp and the human grasp area center, divided by the object's maximum extent; $\tilde{s}_{feas} \in [0, 1]$ is the grasp feasibility score which quantifies robot-specific feasibility (see Sec. 4.2). The weights $(w_{heat}, w_{dist}, w_{feas})$ control the trade-off between these criteria, the grasp with the lowest score being the best compromise.

# 4 Experimental Results

In this section, we first introduce the datasets used for training and evaluation. We give key implementation details and show experimental results. We analyze the performance of the affordance transfer network and compare our framework to state-of-the-art TOH algorithms. Finally, we carry out a user study on a fixed manipulator, and show validation experiments on our legged manipulator.

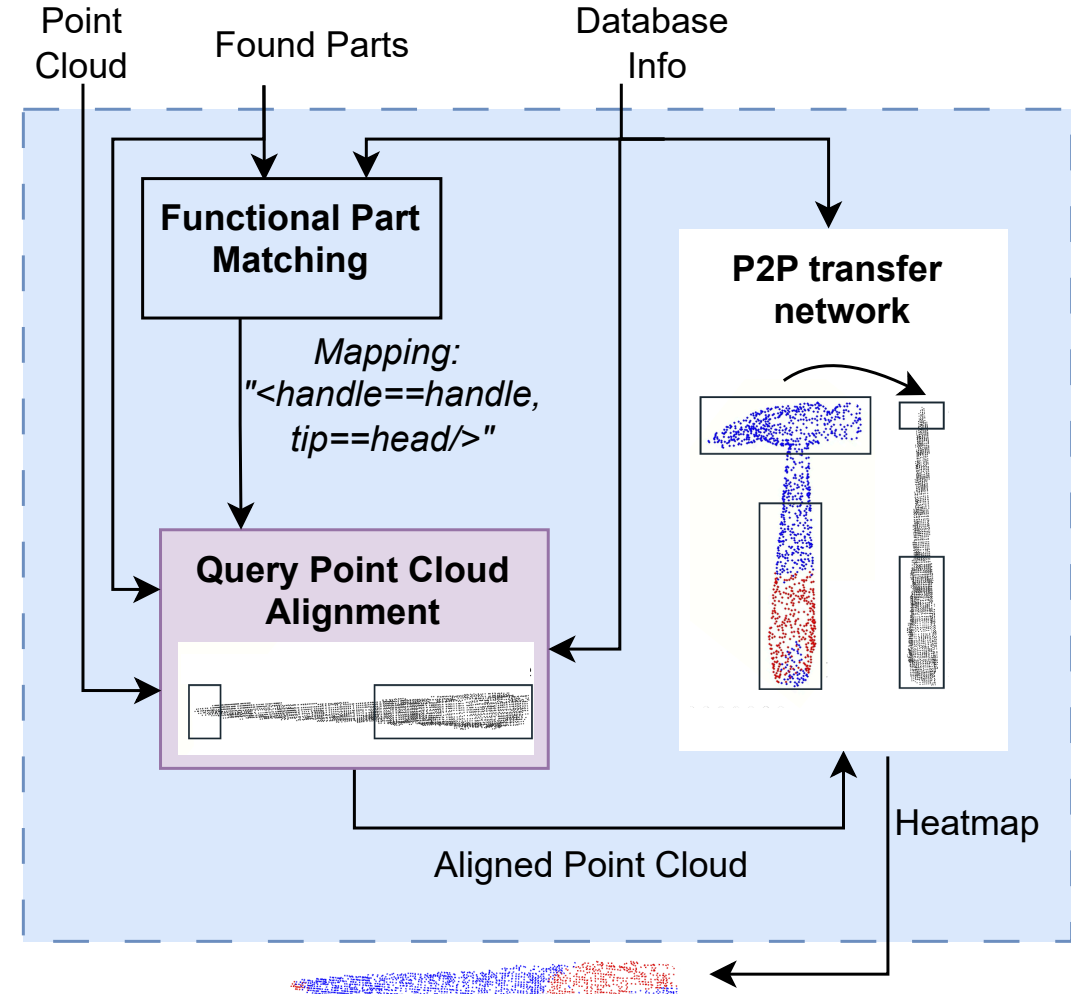

Figure 3: **Affordance Transfer Module Overview.** Inputs are the query object point cloud, database information, and the found parts with their 3D centers. It outputs the transferred affordance upon the query point cloud. Existing components are white (●), proposed components are colored: blue using LLMs (●), purple (●) without.

## 4.1 Datasets

We use three datasets serving separate purposes: network training, affordance transfer evaluation, and comparison to state-of-the-art.

*Training dataset for the affordance transfer network*: We train the point-to-point texture transfer network [6] on meshes of eight household object classes: mug, hammer, pan, knife, screwdriver, scissors, spoon, flashlight (225 meshes in total). We apply geometric augmentations (random rotations and skewing) to improve the descriptor robustness to rotation and mild shape variation. Only meshes are used for training because high-quality object models are more readily available than high-quality real-world scans.

*Evaluation dataset for controlled transfer on partial point clouds*: In contrast to the complete reference models stored in the database, our evaluation uses partial point clouds obtained from single depth-camera captures. This mirrors real robot operation, where typically only partial views are available. Therefore, evaluating affordance transfer on partial observations provides a realistic measure of system performance in handover scenarios. We create a test set of partial point clouds for the eight classes defined above. To generate ground-truth affordances for the partial point clouds, each class is assigned a ground truth reference point cloud from ContactDB [5] or the HANDAL datasets [11] (for spoon and screwdriver). We align the partial cloud to the class reference, propagate the affordance to the partial geometry, and threshold the heatmap into binary human-grasp vs. free regions. Following this procedure, each class yields eight partial object point clouds with annotated baseline affordances. Unless otherwise stated, all quantitative results in Sec. 4.3 use this evaluation dataset. We create both an aligned and rotated dataset, for the latter rotating the partial point clouds by a random axis and angle (up to 360°) after the ground truth transfer.

*State-of-the-art comparison dataset*: For the TOH state-of-the-art comparison, we use the dataset from [36], which contains conventional simple object-task pairs (e.g., knife-cut, mug-drink), conventional complex pairs (e.g., scissor-cut, toothbrush-brush teeth), and unconventional pairs (e.g., screwdriver-hammer, spoon-open a jar).

## 4.2 Implementation Details

For the state-of-the-art comparison and real-world experiments, we use a compact database containing four representative object-task pairs: mug-drink, hammer-hammer, pan-cook, and bottle-drink. This compact database is an intentional design choice to highlight the high degree of generalization achieved from a minimal set. The framework naturally scales with larger and more diverse object libraries, since adding new objects does not require retraining.

The grasp feasibility score differs by platform: on the fixed Franka Panda manipulator, feasibility is defined as the MoveIt! planner's reachability cost [12], while on the legged manipulator, it is defined via the angular deviation between the grasp candidate and the current end-effector pose. The weights for the Grasp Selection module are $w_{heat}$=1.0, $w_{dist}$=0.05 and $w_{feas}$=1.0.

Following [6], we train the affordance transfer network (4 diffusion blocks, 128 output channels) using Adam ($lr = 1e-4$) [19] and self-supervised, alignment, and contrastive losses. We use the first 200 LBO eigenfunctions and Sinkhorn normalization (10 iterations, $\lambda_{sink} = 0.2$), with batch size 32 (multi-class) or 16 (single-class). Hyperparameters are chosen for robustness over partial point clouds.

Each LLM prompt in our framework includes: (i) a text description of the problem and available data, (ii) a structured query specifying the required outputs, and (iii) explicit formatting instructions. Example prompts are provided in the supplementary material.

## 4.3 Affordance Transfer Evaluation

In this section, we first analyze how the affordance transfer network behaves under realistic sensing conditions, then evaluate the generalization and cross-class transferability. In practice, objects are captured as partial point clouds and may appear arbitrarily rotated in the camera view. Without explicit normalization, these factors break descriptor consistency and lead to large performance drops, as described in Sec. 3.2.3. The following ablation quantifies the effect of rotation and validates why normalization is necessary for reliable affordance transfer. The evaluation metrics are: accuracy (fraction of correctly labelled points), precision (fraction of predicted human-grasp points that are correct), and recall (fraction of ground-truth human-grasp points correctly identified).

Table 1: **Ablation on rotation sensitivity** of the affordance transfer network (mug class). We report Accuracy, Recall, and Precision on aligned (baseline) and randomly rotated partial point clouds.

| Feature type | Augmented Rotation (deg) | Accuracy (%) | Recall (%) | Precision (%) |
|---|---|---|---|---|
| **baseline** | 0-30 | **94.64** | **78.60** | **68.77** |
| xyz | 0 | 86.58 | 16.61 | 18.30 |
| | 0-30 | 88.91 | 7.22 | 14.58 |
| | 0-180 | 84.13 | 43.48 | 28.11 |
| | 0-360 | 90.68 | 21.62 | 46.00 |
| hks | 0-360 | 88.12 | 33.22 | 31.61 |
| wks | 0-360 | 91.04 | 50.81 | 48.47 |

### 4.3.1 *Rotation Sensitivity.*

We evaluate the sensitivity of the affordance transfer network to object rotations, since its learned descriptors are not inherently SE(3)-invariant. This analysis is done solely on the mug class. We argue that if the network cannot learn rotation-invariant descriptors for a single class, it is unlikely to generalize reliably across classes. We consider multiple ranges of random rotation magnitudes and sample sizes for this ablation.

Tab. 1 compares a baseline evaluated on the aligned dataset against the randomly rotated dataset introduced in Sec. 4.1. The results highlight the strong sensitivity of the affordance transfer network to object orientation. When query and reference point clouds are geometrically approximately aligned (baseline), the network achieves high accuracy, precision, and recall. However, performance drops significantly once random rotations are introduced, confirming that the learned descriptors are not inherently rotation-invariant. Training with augmented rotations partly mitigates this effect, and using rotation-invariant feature descriptors (HKS [31], WKS [4]) further improves robustness. Still, even with these measures, there remains a considerable gap compared to the aligned baseline, underscoring that alignment is essential for reliable affordance transfer. This observation directly motivates the inclusion of the Query Point Cloud Alignment step in our pipeline (Sec. 3.2.3), which normalizes and aligns query objects before transfer.

### 4.3.2 *Generalization to Novel Classes.*

The goal of this analysis is to evaluate whether the affordance transfer network can learn descriptors that generalize beyond the specific object classes seen during training. Evaluation is done on the aligned dataset (Sec. 4.1). As a baseline, we first train a separate "specialized" network for each class and evaluate it only on that class. As shown in Tab. 2, this baseline achieves strong accuracy (row 1). This is to be expected, since the model is optimized directly for the target distribution. In contrast, the multi-class training setup ($n_{cl} > 1$) scales well and shows an improved cross-class generalization.

Table 2: **Ablation study on generalization across training classes.** Accuracy (%) on aligned partial point clouds when training with different numbers of classes ($n_{cl}$). Rows marked "r" indicate no training data augmentation. Seen/Unseen indicate whether the classes were seen during training.

| | $n_{cl}$ | mug | hammer | pan | knife | scissor | spoon | flashlight | screwdriver |
|---|---|---|---|---|---|---|---|---|---|
| Seen | 1 | 94.92 | 86.99 | 93.91 | 84.56 | 68.01 | 73.88 | 85.69 | 89.28 |
| | 4 | 94.56 | 82.75 | 96.65 | 85.76 | 69.10 | 82.86 | 79.91 | 89.02 |
| | 6 | 94.76 | 85.53 | 96.41 | 84.62 | 71.92 | 83.56 | 79.23 | 88.56 |
| | 6r | 93.72 | 80.22 | 94.53 | 83.85 | 67.73 | 80.53 | 80.33 | 84.52 |
| | 8 | 94.28 | 78.88 | 95.87 | 85.25 | 69.77 | 82.34 | 76.40 | 89.92 |
| Unseen | 1 | 91.29 | 76.02 | 95.12 | 84.76 | 66.92 | 75.81 | 78.21 | 84.49 |
| | 4 | 92.78 | 80.98 | 96.94 | 85.05 | 68.52 | 82.25 | 77.70 | 89.08 |
| | 6 | 92.71 | 82.20 | 96.86 | 84.13 | 69.99 | 82.62 | 78.90 | 89.83 |

Table 3: **Cross-class generalization for affordance transfer.** Accuracy (%) when transferring affordances from a ground-truth source class to a target partial point cloud class (cf. best $n_{cl} = 6$). Seen/Unseen indicate whether the classes were seen during training. Rows labeled "baseline" denote class-matched transfer.

| | | **Target** | | | | | | |
|---|---|---|---|---|---|---|---|---|
| | source | knife | scissor | spoon | flashlight | screwdriver | mug | pan |
| **Seen** | baseline | 85.53 | 71.92 | 83.56 | 79.23 | 89.56 | 94.96 | 96.41 |
| | hammer | 80.93 | 70.73 | 84.92 | 74.68 | 87.15 | 80.12 | 91.44 |
| | mug | 64.96 | 65.96 | 76.79 | 59.64 | 64.39 | - | 94.61 |
| | pan | 80.21 | 71.06 | 79.49 | 65.68 | 65.19 | 94.54 | - |
| **Unseen** | baseline | 84.13 | 69.90 | 82.62 | 76.70 | 89.83 | 92.71 | 96.86 |
| | hammer | 82.83 | 68.98 | 82.99 | 71.58 | 87.86 | 75.50 | 91.01 |
| | mug | 63.93 | 63.45 | 76.36 | 54.93 | 64.26 | - | 95.71 |
| | pan | 81.45 | 69.51 | 76.25 | 65.10 | 67.84 | 92.87 | - |

Table 4: **Comparison to state-of-the-art.** Various household object-task pairs are considered. The metric reflects the successful grasp rate (%). (*) shows ground truth affordances are available.

| Model | conventional | | unconventional | | | | |
|---|---|---|---|---|---|---|---|
| | | | Spoon | Screwdriver | | Toothbrush | Average |
| | easy | *complex* | *open jar* | *hammer* | *play* | *push pin* | |
| GraspGPT | 49 | 47 | **58** | 87 | 87 | 40 | 53 |
| LLM-Handover | **97** | 82 | 21 | 87 | 87 | 60 | 79 |
| OS-TOG* | 74 | 70 | 7 | 53 | 53 | 7 | 62 |
| **AFT-Hand/g4o** | 94 | 87 | 21 | 87 | 87 | 60 | 83 |
| **AFT-Hand/g5** | 95 | **90** | **34** | **87** | **93** | **73** | **86** |
| **AFT-Hand/g5*** | **97** | **93** | **93** | **93** | **93** | **93** | **95** |

When training on multiple classes jointly, inter-class performance increases with the number of classes included in training. As shown in Tab. 2, the best results are achieved with six classes, after which additional classes yield no further improvement. This indicates that exposure to a moderate level of shape diversity is sufficient for the network to acquire descriptors that generalize across categories. Importantly, the performance gap between seen and unseen classes remains small, sometimes even negligible, showing that the model learns transferable descriptors rather than overfitting to class-specific geometry. Finally, augmentations such as rotation and skewing further improve robustness, confirming the importance of data diversity for learning generalizable features.

*4.3.3 Cross-class Transfer.* Having shown rotation robustness and inter-class generalization, we now test the ability to transfer affordances across different object classes. For this, we assess zero-shot transfer by applying ground-truth affordances from a source class (e.g. hammer) to a target partial cloud of a different class (e.g. screwdriver). This evaluation is done on the aligned dataset (Sec. 4.1).

Tab. 3 summarizes accuracy across transfer scenarios, comparing the class-matched baseline with cross-class transfers. As expected, the baseline performs best; however, transferring affordances between semantically similar objects yields accuracy close to the baseline. For instance, transferring from hammers to screwdrivers achieves 87.2 %accuracy, only slightly below the screwdriver baseline of 89.6 %. The same can be observed when transferring from pans to scissors. These results suggest that the network effectively exploits structural and functional similarities between object classes. In contrast, transfers between semantically dissimilar objects lead to significant degradation. Transferring from mugs to knives drops accuracy from 85.5 % to 65.0 %. Similar degradation is observed when transferring from mugs to spoons. Such mismatches highlight the network's reliance on functional similarity between classes for successful generalization.

Additionally, Tab. 3 demonstrates that whether the target class was seen during training has a limited effect compared to functional similarity, underscoring that part-level alignment and canonical orientation are the dominant factors affecting performance.

## 4.4 TOH state-of-the-art comparison

We compare our framework to existing baselines, namely GraspGPT [33], LLM-Handover [36], and OS-TOG [15]. As stated in Sec. 4.1, we use the dataset introduced in [36]. GraspGPT [33] and OS-TOG [15] are state-of-the-art TOG frameworks, which have been shown to perform well for handover tasks. OS-TOG [15] also utilizes a database of known affordances and transfers these affordances to objects of the same class; however, it only considers affordances per object and not per object-task pair. LLM-Handover [36] is a recently developed TOH framework based on LLM reasoning, which showed zero-shot generalization to novel task-object pairs. We run our framework with GPT-4o (gpt-4o-2024-11-20) and GPT-5 (gpt-5-2025-08-07). For the comparison to GraspGPT [33] and LLM-Handover [36], we use GPT-4o to ensure fairness.

Tab. 4 reports rate of correct robot grasps across conventional and unconventional object-task pairs. The metric was computed manually, by visually assessing whether the grasp interferes with the post-handover task: 1 no interference, 0 otherwise. First, it can be noted that using knowledge about the post-handover task (LLM-Handover [36] and AFT-Handover variations) improves the overall robot grasp success rate by 15 − 20 %. OS-TOG [15], which relies on one ground-truth affordance annotation per object, achieves a 62 % success rate and has a significant decrease in performance for unconventional tasks, since it only encodes per-object grasp/non-grasp labels rather than task-conditioned affordances.

AFT-Handover outperforms all baselines: with GPT-4o it achieves 83 % average success rate, and with GPT-5 it further improves to 86 %, showing that stronger LLM reasoning enhances cross-task generalization. As shown in Tab. 4, unconventional object-task pairs remain challenging, even if AFT-Hand/g5 outperforms prior work. In this case, the LLM reasoning fails to correctly match functional parts due to atypical tool use. The most prominent example is spoon/open jar, where the LLM consistently expects the human to use the bowl of the metal spoon instead of the handle end. However, for both simple and complex conventional pairs, the LLM reasoning shows no hallucinations or incorrect part matches. When provided with ground-truth affordances (AFT-Hand/g5*), our method reaches a 95 % success rate, showing the upper bound of our framework. All LLM reasoning errors disappear in this case, even for unconventional tasks, since it can simply self-match database entries. The remaining 5 % gap comes from incorrect part localization (Sec. 3.2.2 ii) and limited grasp candidates. The difference between AFT-Hand/g5* and AFT-Hand/g5 in the conventional cases quantifies the $2 - 3$ % average error introduced by cross-class transfer.

## 4.5 Hardware experiments

In this section, we demonstrate integration of AFT-Handover into existing motion planning frameworks on two different robots. We carry out a comparative user study on a Franka Panda arm and validation experiments in real-world settings on a legged manipulator.

*4.5.1 User study.* We conducted a user study with 14 participants (2 female, 12 male), approved by the ETH Zurich ethics committee, to compare AFT-Handover against LLM-Handover [36] (both running GPT-4o). Participants were recruited from the university community via voluntary sign-up. Each participant carried out seven runs of two robot-human handovers. This resulted in 14 paired trials, enabling paired analysis under identical conditions. The seven objects and their corresponding post-handover tasks were predefined, with one being unconventional (screwdriver to hammer).

The object order was randomized, and each handover was followed directly by the predefined post-handover task. A mockup board enabled realistic task execution. To avoid ordering effects, the method sequence (AFT-Handover vs. baseline) was randomized across participants, who were unaware of which method they were interacting with. However, for each participant, the sequence remained consistent across all objects. After each run, they reported their preference (Pf), regrasping (R), and perceived task understanding (U). Preference indicates which method was favored for that object-task pair, with rates shown in Tab. 5. An overall preference per participant was collected after all seven object-task pairs to capture holistic user choice. Understanding (U) reflects whether the participants think the robot "understood" their intended task. Tab. 5 shows the percentage of "yes" responses. Regrasp (R) is self-reported and indicates whether participants regrasped the object before use. Lower values reflect more efficient handovers. Tab. 5 lists the percentage of regrasps per pair.

The experimental setup included a fixed RealSense D-455 RGB-D camera monitoring the tabletop workspace. Motion planning was handled by MoveIt! [12], ensuring collision avoidance with both the robot itself and the static environment. The handover position was fixed to a predefined area known to the participants to ensure compliance with the robot workspace. The orientation was computed by aligning the axis between the selected robot grasp and the predicted human grasp area with the axis of the robot base pointing towards the human. This ensures that the part the human intends to grasp is oriented directly towards them.

Table 5: **User study results.** Comparison between our method and LLM-Handover across multiple object-task pairs and metrics.

| Method | LLM-Handover | | | AFT-Handover | | |
|---|---|---|---|---|---|---|
| Metrics (%) | Pf | R↓ | U | Pf | R↓ | U |
| Bottle *(pour)* | 43 | 28 | 47 | **57** | **7** | **53** |
| Mug *(drink)* | 41 | 21 | 45 | **59** | **0** | **55** |
| Spoon *(stir)* | 39 | 21 | 47 | **61** | **7** | **53** |
| Pan *(cook)* | 47 | 21 | 41 | **53** | **14** | **59** |
| Hammer *(hammer)* | 44 | 21 | 40 | **56** | **14** | **60** |
| Screwdriver *(screw)* | **52** | **0** | **55** | 48 | 7 | 45 |
| Screwdriver *(hammer)* | 37 | 21 | 42 | **63** | **7** | **58** |
| Average (%) | 43 | 19 | 45 | **57** | **8** | **55** |

As shown in Tab. 5, AFT-Handover outperforms the baseline in nearly all object-task pairs. Only for screwdriver to screw the participants favored the baseline across all metrics. Across all tasks, AFT-Handover increased preference, reduced regrasping, and improved perceived task understanding. Statistical tests confirm these trends. The Wilcoxon signed-rank test was used as it considers both the direction and magnitude of paired differences, offering higher sensitivity. Participants preferred AFT-Handover for significantly more object-task pairs ($p = 0.015$) and reported comparable levels of understanding ($p = 0.055$). Regrasps were significantly fewer with AFT-Handover ($p = 0.027$), and McNemar's test further confirmed an overall reduction ($p = 0.035$, $b = 6$, $c = 17$). A majority of participants (71.43 %) chose AFT-Handover as their overall preference (binomial test: $p = 0.088$; mid-$p = 0.058$). Together, these results indicate that AFT-Handover significantly reduces regrasps and is preferred by users, showcasing its suitability for real-world settings.

*4.5.2 Legged manipulator evaluation.* Demonstrating TOH on a legged robot highlights real-world feasibility beyond controlled tabletop setups. Legged platforms enable assistive and field applications but introduce challenges such as: limited onboard compute due to payload and battery constraints, dynamically shifting viewpoints during locomotion, and perception degradation from motion blur and self-occlusions. These factors make transferring algorithms from fixed manipulators to legged systems non-trivial. Therefore, successfully executing TOH on a mobile platform provides evidence that the proposed method generalizes across embodiments and remains robust under realistic operational conditions.

Our legged manipulator consists of an ANYmal base with a Duatic 6-DoF robotic arm on top. We use an Intel RealSense L515 on the manipulator arm for perception. We test five object-task pairs spanning household and industrial contexts: spoon-stir, pan-cook, screwdriver-screw, screwdriver-hammer, and hammer-hammer. Examples of successful handovers are shown in Fig. 1, with the full execution sequence illustrated in Fig. 5. For high-level control of the robot, we use a custom BehaviorTree implementation, which triggers subsequent modules such as AFT-Handover, human (YOLOv8 [37]) and hand detection (HaMeR [28]), as depicted in Fig. 4.

To complete the handover, a suitable handover pose has to be computed. This is done in the Handover Alignment step (orange in

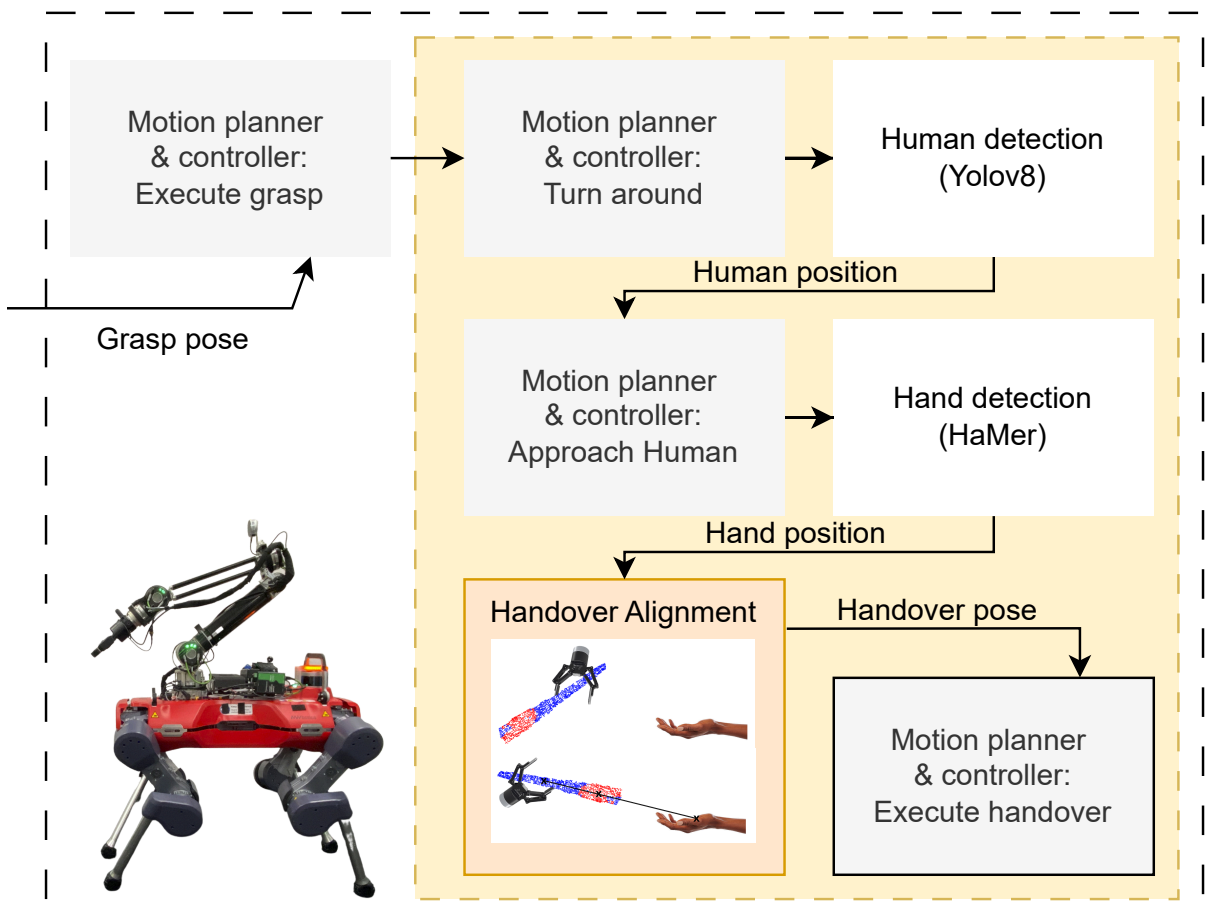

Figure 4: **Handover execution overview on our legged robot**. After picking up the object, the robot turns around, searches for and approaches the human, and detects their hand. From the hand position, we calculate the handover pose (●), which we approach.

Fig. 4). We assume that, after object pickup, the predicted human grasp area from AFT-Handover is rigidly attached to the robot end-effector. The handover position is computed by aligning the center of this region with the hand palm detected by HaMeR[28]. For the orientation, the object is rotated such that the predicted human grasp area directly faces the hand.

We carry out a qualitative evaluation. For each object-task pair, we executed several handovers and assessed whether the object was delivered in a functional orientation and whether the interaction felt natural to the human partner. Whenever the robot successfully grasped the object, the subsequent alignment and handover orientation were correct, and participants could perform the intended task without regrasping. The main difficulties on the legged robot come from less precise arm trajectory tracking and poor grasp candidates, rather than from AFT-Handover itself. This demonstrates that our framework generalizes to mobile legged platforms.

## 5 Limitations

Unconventional tasks remain challenging, with performance drops mainly caused by failures in LLM-based reasoning: mismatching functional parts or inferring implausible task strategies. This limitation reflects the inherent difficulty of reasoning about atypical or ambiguous object use without additional contextual grounding.

Furthermore, our current system does not account for human hand motion during the handover: neither in the user study nor in the legged manipulator experiments. As a result, interactions can feel unnatural, e.g. the robot does not react when the human reaches towards the object. This limitation stems from our single-shot hand detection, a design choice given by the compute constraints of the onboard NVIDIA Jetson Orin rather than conceptual restrictions. Achieving real-time, closed-loop adaptation would greatly improve the interaction, but it introduces a trade-off: lightweight models run fast enough for onboard use but degrade under motion blur or occlusions, while heavier models exceed the allowable inference time. Future work could incorporate reactive tracking through optimized inference pipelines (e.g., TensorRT) or multi-sensor fusion.

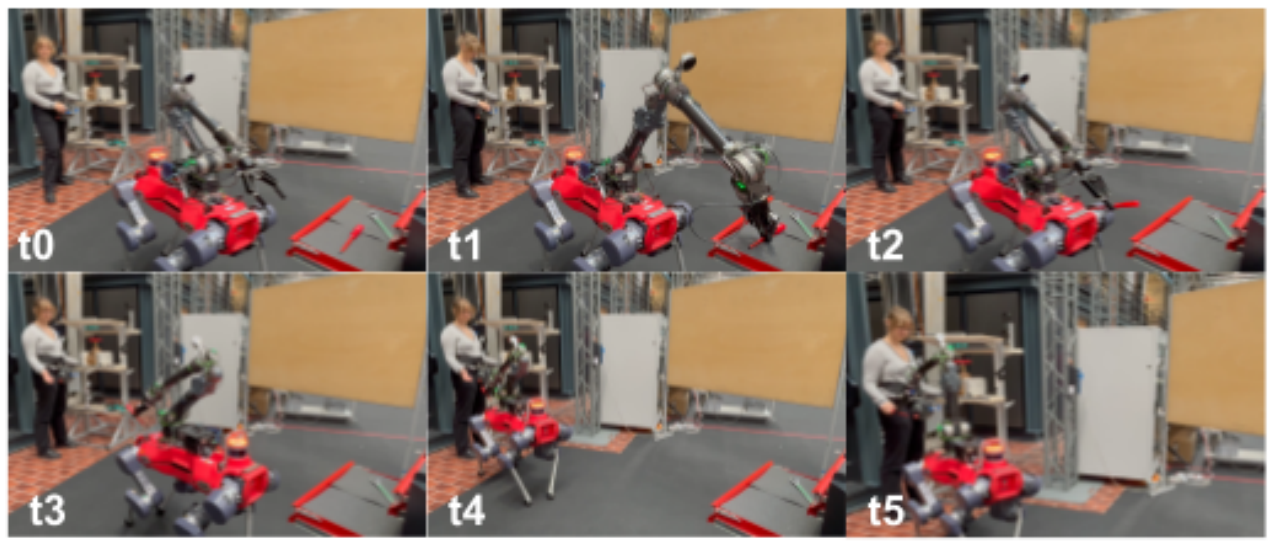

Figure 5: **Handover procedure on our legged manipulator**. [t0: the robot receives the command to bring the red screwdriver for a screwing task; runs AFT-Handover], [t1: the selected grasp is executed, lifting the screwdriver at the shaft], [t2: the robot retracts its arm; starts looking for the human], [t3: the human is detected; approaching starts], [t4: the robot reaches the human; starts scanning for the human hand], [t5: TOH is executed]

From the user study, we observed that humans have individual preferences for how objects should be handed over. Incorporating human feedback during execution could therefore not only correct the occasional errors but also adapt the handover to user-specific preferences. Additionally, for some post-handover tasks, e.g. stirring or screwing, the appropriate handover orientation and the subsequent human grasp depend on how much force is needed for the task (power vs. pinch grasp). Hence, more information about the post-handover scene could improve the interaction quality. Finally, the motion planner is currently not aware of object states, which can lead to issues, e.g. fluid inside a full mug being spilled.

## 6 Conclusions and future work

This paper introduces AFT-Handover, a framework for task-oriented robot-human handovers that combines LLM-guided reasoning with texture-based affordance transfer. Unlike prior work that models affordances in an object- or task-specific way, AFT-Handover leverages a compact database of object-task pairs and establishes part-level correspondences through LLMs, enabling zero-shot affordance transfer across functionally distinct objects.

We demonstrated the effectiveness of our approach in three stages. First, we analyzed the affordance transfer network in isolation, showing its sensitivity to rotation and the necessity of our alignment step, as well as its ability to generalize to unseen object classes and transfer affordances across semantically similar tools. Second, we compared AFT-Handover against state-of-the-art baselines, which we outperform on both conventional and unconventional tasks. Finally, we validated our framework in real-world experiments, including a user study on a fixed manipulator arm and demonstrations on a legged mobile manipulator.

Our results support that AFT-Handover bridges semantic reasoning and affordance transfer to enable generalized TOH. While unconventional tasks remain challenging, we highlight the potential of combining LLM-driven reasoning with affordance transfer for human-centered manipulation. Future work will focus on improving the reactivity and naturalness of the process by dynamically adjusting handover orientation and location in response to human motion and incorporating more post-handover task context.